\documentclass{article}
\usepackage{spconf,amsmath,graphicx}
\usepackage{adjustbox}

\title{Fully complex-valued deep learning model for visual perception}
\name{Aniruddh Sikdar$^{1 \dagger}$ , \quad \quad Sumanth Udupa$^{2 \dagger}$, \quad \quad Suresh Sundaram$^{2}$ \thanks{$^{\dagger}$Equal contribution of authors.}}

\address{Author Affiliation(s)}
\address{ 
$^{1}$Robert Bosch Centre for Cyber Physical Systems, Indian Institute of Science, Bengaluru\\
$^{2}$Department of Aerospace Engineering, Indian Institute of Science, Bengaluru, \\ 
{\tt \{aniruddhss, sumanthudupa, vssuresh\} @iisc.ac.in}
}

\begin{document}
\thispagestyle{empty}
\onecolumn 
\vspace*{\fill}
This work has been submitted at the IEEE International Conference on Acoustics, Speech and Signal Processing 2023. Copyright may be transferred without notice, after which this version may no longer be accessible.
\vspace*{\fill}

\twocolumn
\newpage
\maketitle
\begin{abstract}
Deep learning models operating in the complex domain are used due to their rich representation capacity. However, most of these models are either restricted to the first quadrant of the complex plane or project the complex-valued data into the real domain, causing a loss of information. 
This paper proposes that operating entirely in the complex domain increases the overall performance of complex-valued models.
A novel, fully complex-valued learning scheme is proposed to train a Fully Complex-valued Convolutional Neural Network (FC-CNN) using a newly proposed complex-valued loss function and training strategy. 
Benchmarked on CIFAR-10, SVHN, and CIFAR-100, FC-CNN has a 4-10\% gain compared to its real-valued counterpart, maintaining the model complexity. With fewer parameters, it achieves comparable performance to state-of-the-art complex-valued models on CIFAR-10 and SVHN. For the CIFAR-100 dataset, it achieves state-of-the-art performance with  25\% fewer parameters. FC-CNN shows better training efficiency and much faster convergence than all the other models.
\end{abstract}
\section{Introduction}
\label{sec:intro}
Deep learning is used for various tasks such as speech processing \cite{amodei2016deep}\cite{oord2016wavenet}, robotics, signal processing, image processing, and sonar \cite{bassey2021survey}, which contain complex numbers, but the majority of the models rarely use complex numbers. Complex-valued data can exist in: (1) natural form, involving applications using Synthetic Aperture Radar (SAR) \cite{chen2021cvcmff}, Magnetic Resonance Imaging (MRI), and telecommunication, (2) complex-valued representation of real-valued data, such as Fourier transform \cite{kondor2018generalization} and spectrum-based methods \cite{maire2016affinity}, and (3) physics application such as optical physics \cite{nie2021neural}, where both amplitude and phase information are essential. Complex-valued models (CV-models) have certain advantages over their real-valued counterparts, such as richer representation ability \cite{nitta2003computational}, easier optimization \cite{georgiou1992complex}, efficient multi-task learning \cite{cheung2019superposition}, superior MRI reconstruction \cite{sinha2007parallel}, and robust memory retrieval mechanism \cite{wolter2018complex}. \\
CV models can be used for standard computer vision tasks, such as image classification, because of their rich representation capacity in the complex domain. 
DCN \cite{DBLP:journals/corr/TrabelsiBSSSMRB17} extends real-valued deep learning models to the complex domain by proposing complex-valued batch norm, weight initialization, and CReLU activation function. It operates only in the first quadrant of the complex plane. SurReal \cite{chakraborty2020surreal} tackles complex numbers from a manifold deep learning perspective but does not exploit the complex-valued representation. CDS models \cite{singhal2022co} introduce equivariant and invariant counterparts of common neural network layers to address complex scaling. \\
All the models mentioned above do not operate entirely in the complex domain and project data to the real domain to make predictions and compare with ground truth labels. This causes a loss of phase information, hence, the complex domain's representation capacity is not used optimally. \cite{nie2021neural} shows that more information is stored in the phase component compared to the amplitude component in the parameter subspace, proving the heterogeneous nature of neural networks. Operating completely in the complex domain to train the model may improve the overall generalization ability and performance of CV models for visual perception.\\
In this paper, a  Fully Complex-valued Convolutional Neural Network (FC-CNN) is proposed for image classification with the same number of parameters as
its real-valued counterpart. A novel complex-valued learning scheme is proposed to train FC-CNN entirely in the complex domain, taking phase information into account when updating network parameters. The key contributions can be summarized as follows: 
(1) A novel fully complex-valued learning scheme is proposed using orthogonal decision boundary theory \cite{nitta2004orthogonality} to directly convert shallow real-valued networks to fully complex-valued networks, which improves the performance while maintaining the same model complexity. (2) A new regularized complex-valued loss function and a two-step training strategy for FC-CNN are proposed to address overfitting, leading to faster learning and better convergence. (3) FC-CNN outperforms its real-valued counterpart on RGB and other encodings. It achieves state-of-the-art performance on CIFAR-10, CIFAR-100, and SVHN datasets, with less parameters.
\section{Fully complex-valued convolutional neural network}
In this section, the Fully Complex-valued Convolutional Neural Network (FC-CNN) is explained in detail. Complex-valued prerequisites are briefly discussed first, followed by network architecture and the complex-valued learning scheme.
\subsection{Complex-valued Prerequisites}
\textbf{Complex-valued convolution operation}
Complex-valued convolution operator is defined for complex-valued weight $W$ and complex-valued input $I$ as follows,
\begin{equation}
  \begin{aligned}
    \textbf{W} \ast \textbf{I} & = ( W_R +\emph{i}W_I) \ast ( I_R +\emph{i}I_I) \\ 
  \end{aligned}
\end{equation}
where $W_R$ ,$W_I$ denote the real and the imaginary parts of the complex-valued weight, and $I_R$ ,$I_I$ denote the real and imaginary parts of the complex-valued input. \\
\textbf{Complex-valued activation function} 
Complex differentiable functions are called analytic or holomorphic if the real and imaginary parts of the complex function follow Cauchy-Riemann differential equations. A sufficient condition for complex-valued activation functions to be compatible with complex backpropagation \cite{georgiou1992complex} is that they should be differentiable with respect to each complex-valued parameter and its complex conjugate \cite{kreutz2009complex}. Although there are numerous complex-valued activation functions, mainly CReLU \cite{DBLP:journals/corr/TrabelsiBSSSMRB17} and complex cardioid \cite{virtue2019complex} activation functions are considered as both are extensions of the ReLU function in the complex plane. CReLU can be defined as applying the ReLU function on  real and imaginary parts separately, as shown below,
\begin{equation}
  \begin{aligned}
      CRelu(z) & = ReLU(R(z)) + i \cdot ReLU(Img(z)), \\
  \end{aligned}
\end{equation}
where z is a complex number, R(z) and Img(z) represent the real and imaginary components of the complex number respectively. While using CReLU, the complex plain is restricted to only the first quadrant.
The complex cardioid is defined as, 
\begin{equation}
  \begin{aligned}
      f(z) & =\frac{1}{2}(1+cos(\angle z))\cdot z\\
  \end{aligned}
\end{equation}
where z is a complex number, $\angle$z denotes the phase of the complex number. The cardioid function operates in the whole complex-plain while preserving phase and scaling the magnitude.
\subsection{Fully complex-valued learning scheme}
\begin{figure}[ht]
    \centering
    \includegraphics[width=8.8cm]{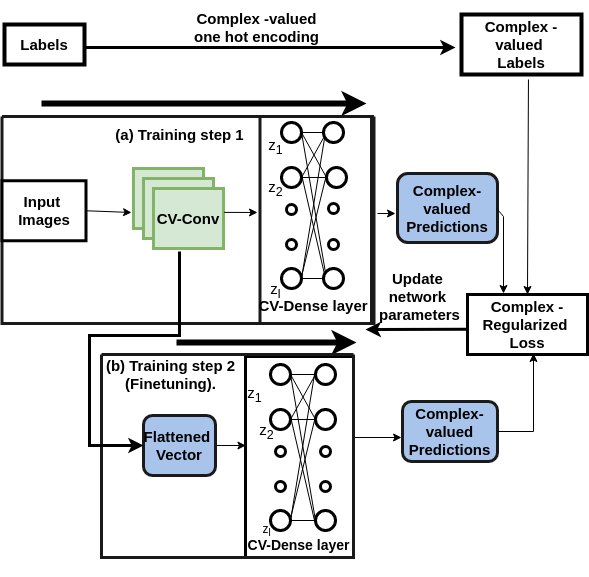}
    \caption{Fully complex-valued learning scheme with two stage training strategy.}
    \label{fig:CV-train}
\end{figure}
\textbf{Network architecture}
FC-CNN model has 3 convolutional layers (stride 2) and 1 fully connected layer, as shown in Table 1. The group setting for the second and third convolutional layers is 2 and 4, respectively. Depthwise-separable convolution is used as a learnable pooling layer to reduce the feature map size, with a group setting of 64.
The real and imaginary components of the complex-valued weights are initialized using a uniform distribution. Cardioid activation function is used to operate in the whole complex plane and not be restricted to any particular quadrant. The output of the complex-valued fully connected layer is a complex-valued entity \cite{Suresh2013-2}. The last layer is modified according to the number of classes for a given dataset. \\
\textbf{Complex-valued one hot encoding} Real-valued labels are one hot encoded to complex-valued labels to compare with the complex-valued predictions made by the fully complex-valued network. Let $\{(x_1,c_1),(x_2,c_2)....(x_t,c_t) ... (x_n,c_n) \}$ be the input  data, where $x_t$ represents the t-th input image and $c_t$ represents its label\cite{Suresh2013}\cite{suresh2013supervised}. One-hot encoding $y_k^t$ =$\{y_1^t,y_2^t,..y_k^t...y_n^t\}$ for its corresponding label $c_t$, is given by :
\begin{equation*}
y_k^t = \left\{
        \begin{array}{ll}
            1 + i, & \text{if} \quad c_t=k \\
            -1 -i, & \quad \text{otherwise.}
        \end{array}
    \right.
\end{equation*}

\begin{table}[]
 \setlength{\arrayrulewidth}{1pt}
\begin{adjustbox}{max width=0.48\textwidth} 
\renewcommand{\arraystretch}{1.2}
\begin{tabular}{|c|c|c|c|c|}
\hline
Layer Type                                                    & Input shape                        & Kernel                 & Stride & Output shape                   \\ \hline
CV - Conv1                                                    & {[}3,32,32{]}                      & 3x3                    & 2      & {[}32,16,16{]}                 \\ \hline
Cardioid                                                      & {[}32,16,16{]}                     & -                      & -      & {[}32,16,16{]}                 \\ \hline
CV - Conv2                                                    & {[}32,16,16{]}                     & 3x3                    & 2      & {[}64,8,8{]}                   \\ \hline
Cardioid                                                      & {[}64,8,8{]}                       & -                      & -      & {[}64,8,8{]}                   \\ \hline
CV - Conv3                                                    & {[}64,8,8{]}                       & 3x3                    & 2      & {[}64,4,4{]}                   \\ \hline
Cardioid                                                      & {[}64,4,4{]}                       & -                      & -      & {[}64,4,4{]}                   \\ \hline
\begin{tabular}[c]{@{}c@{}}CV- \\ Depthwise Conv\end{tabular} & {[}64,4,4{]}                       & 4x4                    & 2      & {[}128,1,1{]}                  \\ \hline
\multicolumn{1}{|c|}{Flatten Layer}                           & \multicolumn{1}{c|}{{[}128,1,1{]}} & \multicolumn{1}{c|}{-} & -      & \multicolumn{1}{c|}{{[}128{]}} \\ \hline
\begin{tabular}[c]{@{}c@{}}CV - Linear \\ layer\end{tabular}  & {[}128{]}                          & -                      & -      & {[}10{]}                       \\ \hline
\end{tabular}
\end{adjustbox}
\caption{FC-CNN model architecture for 10 output classes.}
\end{table}

\begin{table*}[]
 \setlength{\arrayrulewidth}{1pt}
\begin{adjustbox}{max width=0.98\textwidth} 
\renewcommand{\arraystretch}{1.2}
\begin{tabular}{ccccccccccc}
                                                                                       &                                     &                            &                               &                              &                            &                            &                              &                            &                                &                              \\ \hline
\multicolumn{1}{|c|}{Method}                                                           & \multicolumn{1}{c|}{\# Param}       & \multicolumn{1}{c|}{}      & \multicolumn{1}{c|}{CIFAR-10} & \multicolumn{1}{c|}{}        & \multicolumn{1}{c|}{}      & \multicolumn{1}{c|}{SVHN}  & \multicolumn{1}{c|}{}        & \multicolumn{1}{c|}{}      & \multicolumn{1}{c|}{CIFAR-100} & \multicolumn{1}{c|}{}        \\ \hline
\multicolumn{1}{|c|}{}                                                                 & \multicolumn{1}{c|}{}               & \multicolumn{1}{c|}{RGB}   & \multicolumn{1}{c|}{LAB}      & \multicolumn{1}{c|}{Sliding} & \multicolumn{1}{c|}{RGB}   & \multicolumn{1}{c|}{LAB}   & \multicolumn{1}{c|}{Sliding} & \multicolumn{1}{c|}{RGB}   & \multicolumn{1}{c|}{LAB}       & \multicolumn{1}{c|}{Sliding} \\ \hline
\multicolumn{1}{|c|}{Real-valued CNN}                                                  & \multicolumn{1}{c|}{22,634/ 34,244} & \multicolumn{1}{c|}{67.27}      & \multicolumn{1}{c|}{62.52}         & \multicolumn{1}{c|}{65.0}        & \multicolumn{1}{c|}{87.3}      & \multicolumn{1}{c|}{86.58}      & \multicolumn{1}{c|}{87.11}        & \multicolumn{1}{c|}{37.81}      & \multicolumn{1}{c|}{33.49}          & \multicolumn{1}{c|}{36.48}        \\ \hline
\multicolumn{1}{|c|}{DCN}                                                              & \multicolumn{1}{c|}{23,914/ 47,044} & \multicolumn{1}{c|}{65.64}      & \multicolumn{1}{c|}{63.72}         & \multicolumn{1}{c|}{66.94}        & \multicolumn{1}{c|}{86.80}      & \multicolumn{1}{c|}{85.77}      & \multicolumn{1}{c|}{86.56}        & \multicolumn{1}{c|}{35.88}      & \multicolumn{1}{c|}{35.25}          & \multicolumn{1}{c|}{34.85}        \\ \hline
\multicolumn{1}{|c|}{Sur-Real}                                                         & \multicolumn{1}{c|}{35,274/-}       & \multicolumn{1}{c|}{50.68} & \multicolumn{1}{c|}{53.02}    & \multicolumn{1}{c|}{54.61}   & \multicolumn{1}{c|}{80.51} & \multicolumn{1}{c|}{53.48} & \multicolumn{1}{c|}{80.79}   & \multicolumn{1}{c|}{23.57} & \multicolumn{1}{c|}{25.97}     & \multicolumn{1}{c|}{26.66}   \\ \hline
\multicolumn{1}{|c|}{CDS Type - I}                                                     & \multicolumn{1}{c|}{24,241/ 47281}  & \multicolumn{1}{c|}{69.18} & \multicolumn{1}{c|}{67.89}    & \multicolumn{1}{c|}{\textbf{70.28}}   & \multicolumn{1}{c|}{89.40} & \multicolumn{1}{c|}{\textbf{89.27}} & \multicolumn{1}{c|}{89.81}   & \multicolumn{1}{c|}{40.2}  & \multicolumn{1}{c|}{37.94}     & \multicolumn{1}{c|}{39.27}   \\ \hline
\multicolumn{1}{|c|}{CDS Type - E}                                                     & \multicolumn{1}{c|}{23,697/ 46737}  & \multicolumn{1}{c|}{\textbf{69.98}} & \multicolumn{1}{c|}{67.7}     & \multicolumn{1}{c|}{69.5}    & \multicolumn{1}{c|}{88.75} & \multicolumn{1}{c|}{74.38} & \multicolumn{1}{c|}{\textbf{90.50}}   & \multicolumn{1}{c|}{40.11} & \multicolumn{1}{c|}{38.91}     & \multicolumn{1}{c|}{40.97}   \\ \hline
\multicolumn{1}{|c|}{\begin{tabular}[c]{@{}c@{}}FC-CNN\end{tabular}} & \multicolumn{1}{c|}{\textbf{22,634/ 34,244}} & \multicolumn{1}{c|}{69.82}      & \multicolumn{1}{c|}{\textbf{69.87}}         & \multicolumn{1}{c|}{69.99}        & \multicolumn{1}{c|}{\textbf{89.65}}      & \multicolumn{1}{c|}{89.13}      & \multicolumn{1}{c|}{89.12}        & \multicolumn{1}{c|}{\textbf{41.71}}      & \multicolumn{1}{c|}{\textbf{42.26}}          & \multicolumn{1}{c|}{\textbf{41.46}}        \\ \hline
\end{tabular}
\end{adjustbox}
\caption{Comparison of test accuracy between real-valued CNN, state of the art complex-valued and fully complex-valued model (FC-CNN) on CIFAR-10, CIFAR-100 and SVHN.}
\end{table*}
\textbf{Orthogonal decision boundary theory}
A fully complex-valued network consists of two hypersurfaces as its decision boundary - real and imaginary \cite{nitta2004orthogonality}. These hypersurfaces are orthogonal to each other if the activation function in the network satisfies Cauchy–Riemann equations (lemma 6.1, \cite{Suresh2013}). Real and imaginary hypersurfaces of a fully complex network with complex cardioid are orthogonal to each other as complex cardioid satisfies Cauchy–Riemann equations \cite{virtue2019complex}. Complex-valued one hot encoding of the ground truth labels are projected in the output feature space. Either the real or imaginary hypersurface can be used to classify the on and off values of the labels.
Using orthogonal decision boundary theory, the real components of the complex-valued output can be directly compared with the real components of the complex-valued one-hot encoded labels to compute the accuracy.\\
\textbf{Regularized complex-valued hinge loss} A new loss function is proposed to compute the error between complex-valued predictions and complex-valued labels. This helps the phase information to be taken into account when updating the network parameters.
The complex-valued hinge loss \textbf{e} \cite{suresh2013supervised} is given by , 
\begin{equation*}
e = \left\{
        \begin{array}{ll}
            
            0 & \text{if} \quad Re(y_l^t) \cdot Re (\hat y_l^t) >  1 \\
            (y_l^t)-(\hat y_l^t) & \quad \quad  \text{otherwise.}
            
        \end{array}
    \right.
\end{equation*}
where $(y_l^t)$ is the complex-valued one hot encoding and $(\hat y_l^t)$ is the complex-valued predictions. FC-CNN has an overfitting issue, which is why the loss function is modified to have a regularizing effect for the correctly predicted samples. Two terms, error threshold $e_{thr}$ and max threshold $e_{M}$ are defined, where error threshold $e_{thr}$ is updated after every epoch and is given by, 
\begin{equation}
  \begin{aligned}
      e_{thr} & = e^{-(0.05) \cdot (epoch)}\\
  \end{aligned}
\end{equation}
Maximum threshold $e_{M}$ represents the maximum value of error from all the samples in the batch, and is defined as, 
\begin{equation}
  \begin{aligned}
      e_{M} & = \text{max}( \text{abs} (e(y_l^t,\hat y_l^t))) \\
  \end{aligned}
\end{equation}
where \textit{abs} represents the magnitude of complex numbers and max operator returns the maximum value from all the samples in the batch. The following condition is used to update \textit{e},
\begin{equation*}
e = \left\{
        \begin{array}{ll}
            
            0 & \text{if} \quad e_{M} < e_{thr} \\
            e & \text{if} \quad e_{M} > e_{thr}
            
        \end{array}
    \right.
\end{equation*}
For misclassified samples, the complex-valued hinge loss \textit{e} remains the same.
Real-valued loss function E is defined as the product of complex-valued hinge loss \textbf{e} and its complex conjugate\cite{Suresh2013-2},and is given by\\
\begin{equation}
  \begin{aligned}
      E & =\frac{1}{2n} (e^H)(e)\\
  \end{aligned}
\end{equation}
where n is the number of data samples. \\
\textbf{Training strategy} The fully complex-valued network (FC-CNN) is trained end-to-end using the proposed complex loss function and complex-valued labels, as shown in Fig.\ref{fig:CV-train}(a) training step 1. Once the whole network is trained, the output of the final complex-valued convolutional layer of the last epoch is flattened and stored. The model is then finetuned in the second training step (Fig.\ref{fig:CV-train}(b)), where only the complex-valued linear layer is re-trained. During this step, the stored flattened vector is fed to the complex-valued linear layer and trained for the same number of epochs with the same loss function. During inference time, the model consists of complex-valued convolutional layers and the finetuned complex-valued linear layer.
\vspace{3mm}
\begin{figure}[ht]
    \centering
    \includegraphics[width=9.5cm]{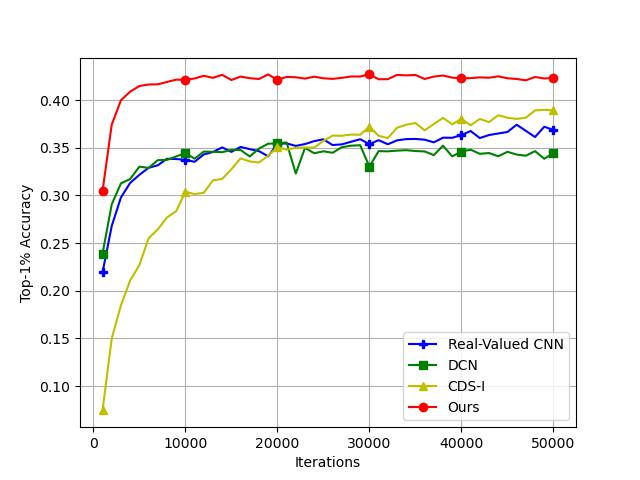}
    \caption{Performance curves of test accuracy of real-valued CNN, CV-models and FC-CNN on CIFAR-100 (RGB) dataset.}
    \label{fig:curves}
\end{figure}
\begin{table}[]
\begin{tabular}{|c|c|c|}
\hline
Method                                                          & \begin{tabular}[c]{@{}c@{}}CIFAR-100\\    (RGB)\end{tabular} & \begin{tabular}[c]{@{}c@{}}FLOPs\\ (MACS)\end{tabular} \\ \hline
\begin{tabular}[c]{@{}c@{}}Real -valued  CNN\end{tabular}     & 37.81                                                       & 1M                                                     \\ \hline
\begin{tabular}[c]{@{}c@{}}DCN  (With CReLU)\end{tabular}     & 35.88                                                        & 1.03M                                                  \\ \hline
\begin{tabular}[c]{@{}c@{}}DCN  (With Cardioid)\end{tabular}  & 34.29                                                        & 999.65K                                                \\ \hline
\begin{tabular}[c]{@{}c@{}}FC-CNN (W/o Finetuning)\end{tabular}  & 39.06                                                              & 986.85K                                                \\ \hline
\begin{tabular}[c]{@{}c@{}}FC-CNN (With Finetuning)\end{tabular} & \textbf{41.71}                                                        & \textbf{986.85K}                                                \\ \hline
\end{tabular}
\caption{Comparison of test accuracy and FLOPs (forward pass) of real-valued, DCN and fully complex-valued (FC-CNN) models. FLOPs are reported during testing.}
\end{table}
\section{Experimental results}
\label{sec:typestyle}
The experiments are conducted on three benchmark datasets, i.e., CIFAR-10 \cite{krizhevsky2009learning}, CIFAR-100, and SVHN \cite{netzer2011reading}. CIFAR-10 and CIFAR-100 contain 10 and 100 classes, with 50000 training images and 10000 testing images, respectively. SVHN contains ten classes, with 73,257 training images and 26,032 test images. For a fair comparison, real-valued CNN and DCN both have the same network architecture as FC-CNN and are trained using cross-entropy loss. Relu activation function is used for the real model, and CReLU activation is used for the DCN model.
Sur-real \cite{chakraborty2020surreal}, CDS Type - I, and E models \cite{singhal2022co} have the same network architecture as mentioned in \cite{singhal2022co}. Note that as Sur-real's code is not publicly available, the results mentioned in Table 2 were borrowed from the results cited in \cite{singhal2022co}.\footnote[1]{The results for CDS-I and E are reproduced from the github code : https://github.com/sutkarsh/cds. The results for FC-CNN are reproduced using the same code and same seed.}
All the models are trained with AdamW optimizer using momentum (0.99,0.999), with a learning rate of 0.001, batch size 256, and weight decay of 0.1.\\
Test accuracy of real-valued and complex-valued models are shown in Table 2 for real-valued RGB  images and their corresponding complex-valued sliding \cite{singhal2022co} and LAB encoding \cite{singhal2022co} for all three datasets. The number of parameters for all the models drastically increases when changing the output categories from 10 to 100. The real-valued CNN and DCN models have comparable performance to each other, but for RGB data, real-valued CNN outperforms the DCN model. Comparing these networks to fully complex-valued network(FC-CNN), a significant improvement in the performance can be seen. For CIFAR-10 and SVHN datasets, FC-CNN achieves similar performance to the best-performing CDS model, with fewer parameters. 
For the CIFAR-100 dataset, FC-CNN outperforms its real-valued counterpart on RGB and all the encodings with the same number of parameters. It achieves state-of-the-art performance compared to other CV models with considerably less number of parameters. Fig.\ref{fig:curves} shows the test accuracy of real-valued, complex-valued, and FC-CNN models on CIFAR-100 (RGB). FC-CNN outperforms all the other models with much more efficient learning. It converges within 15k iterations, while all the other models take considerably more training time. Compared to its real-valued counterpart, with the same number of parameters, the overall test accuracy and convergence are much better. Thus, it generalizes better on larger datasets than real-valued and CV models. \\
Table 3 shows the test accuracy and model complexity of real-valued, DCN, and FC-CNN networks on the CIFAR-100 dataset. FLOPs for the real-valued and FC-CNN networks remain almost the same. FC-CNN performs better than both networks without the finetuning step.
Using either CReLU or cardioid for the DCN model does not improve its performance compared to the real-valued network. Hence, we infer that operating in all quadrants of the complex plane is not enough, and the model must also train entirely in the complex domain and not be projected back to the real domain. The empirical results of FC-CNN justify this intuition. The two-stage training strategy, i.e., finetuning the CV-linear layer, shows a considerable gain in performance.

\section{Conclusion}
\label{sec:majhead}
This paper proposes a Fully Complex-valued network for image classification of real-valued images and their complex-valued encodings. It is an extension of a shallow real-valued CNN to the complex domain with an equal number of parameters. A novel, fully complex-valued learning scheme is proposed for the FC-CNN to operate entirely in the complex domain and operate in all four quadrants of the complex domain, using a novel regularized loss function and training strategy. Experimental results on CIFAR-10, SVHN, and CIFAR-100 show that FC-CNN has a gain in accuracy of 4-10\% compared to its real-valued counterpart on RGB, Sliding, and LAB encodings. Compared to the shallow DCN model, accuracy increases by 4-6\% with 5\% lesser computational complexity. Compared to other complex-valued models, it achieves comparable performance to the state-of-the-art network on CIFAR-10, SVHN. For CIFAR-100, it achieves state-of-the-art performance with almost 25\% fewer parameters.

\bibliographystyle{IEEEbib}
\bibliography{egbib,refs}

\end{document}